\def\doctitle{Automatic localization and decoding of honeybee markers using deep convolutional neural networks}
\def\docauthor{Benjamin Wild}

\documentclass[12pt,a4paper]{scrartcl}

\usepackage[utf8]{inputenc}

\usepackage{fancyref}

\usepackage{scalerel}
\newcommand\equalhat{%
\let\savearraystretch\arraystretch
\renewcommand\arraystretch{0.3}
\begin{array}{c}
\stretchto{
    \scalerel*[\widthof{=}]{\wedge}
    {\rule{1ex}{3ex}}%
}{0.5ex}\\
=%
\end{array}
\let\arraystretch\savearraystretch
}

\usepackage{amsmath}

\usepackage[T1]{fontenc}
\usepackage{libertine} \usepackage[libertine]{newtxmath}

\addtokomafont{disposition}{\rmfamily}

\addtokomafont{disposition}{\rmfamily}

\usepackage[format=plain,justification=justified]{caption}
\usepackage{fancyhdr}
\usepackage{xcolor}
\usepackage{url}
\usepackage{makeidx}
\usepackage{listings}
\usepackage{mdframed}
\usepackage{xparse}
\usepackage[export]{adjustbox}
\usepackage{realboxes}
\usepackage{calc}
\usepackage{fp}
\usepackage{tikz}
\usepackage{pgfplots}
\usepackage{todonotes}
\usepackage{booktabs}
\usepackage{cleveref}
\usepackage{graphicx}
\usepackage{color}
\usepackage{tikz}

\usepackage{chngcntr}
\counterwithin{figure}{section}

\usepackage{setspace}
\usepackage[status=draft,layout=margin]{fixme}

\usepackage{natbib}

\fxsetup{marginface=\singlespacing}

\pgfplotsset{compat=newest}
\pgfplotsset{plot coordinates/math parser=false}

\usetikzlibrary{backgrounds}
\usetikzlibrary{arrows,positioning}
\tikzset{
    >=stealth',
    punkt/.style={
           rectangle,
           rounded corners,
           draw=black, very thick,
           text width=6.5em,
           minimum height=2em,
           text centered},
    pil/.style={
           ->,
           thick,
           shorten <=2pt,
           shorten >=2pt,}
}

\usepackage{hyperref}
\hypersetup{%
    pdfencoding=auto,
    pdfauthor={\docauthor},
    pdftitle={\doctitle},
    colorlinks=false,
    pdfborder={0 0 0}
}

\definecolor{dkgreen}{rgb}{0,0.6,0}
\definecolor{gray}{rgb}{0.5,0.5,0.5}
\definecolor{mauve}{rgb}{0.58,0,0.82}
\definecolor{listingbg}{rgb}{0.96,0.96,0.96}

\NewDocumentEnvironment{docfigure}{m O{defaultlabel}}{
	\begin{figure}[htb!]
		\begin{mdframed}[linewidth=0pt,backgroundcolor=listingbg, align]
            \begin{center}
				}{
            \end{center}
			\caption{#1}\label{fig:#2}
		\end{mdframed}
	\end{figure}
}

\newcommand{\docgraph}[2]{
    \def\res{0.0}
	\FPdiv{\res}{1}{#2}
	\def\ressub{0.0}
	\FPmul{\ressub}{\res}{0.05}
	\includegraphics[width=\res\textwidth - \ressub\textwidth]{img/#1}
}

\newlength\figurewidth

\newcommand\ie{i.\,e.\ }
\newcommand\eg{e.\,g.\ }

\DeclareOldFontCommand{\bf}{\normalfont\bfseries}{\mathbf}

\begin{document}

\begin{flushleft}
{\Large
\textbf\newline{\doctitle}
}
\newline
\\
Benjamin Wild\textsuperscript{1}, 
Leon Sixt\textsuperscript{1},
Tim Landgraf\textsuperscript{1, *}
\\
\bigskip
\bf{1} Dahlem Center of Machine Learning and Robotics, Freie Universit{\"a}t Berlin
\\
\bigskip
* tim.landgraf@fu-berlin.de 

\end{flushleft}

\section*{Abstract}
The honeybee is a fascinating model animal to investigate how collective behavior emerges from (inter-)actions of thousands of individuals. Bees may acquire unique memories throughout their lives. These experiences affect social interactions even over large time frames. Tracking and identifying all bees in the colony over their lifetimes therefore may likely shed light on the interplay of individual differences and colony behavior. This paper proposes a software pipeline based on two deep convolutional neural networks for the localization and decoding of custom binary markers that honeybees carry from their first to the last day in their life. We show that this approach outperforms similar systems proposed in recent literature. By opening this software for the public, we hope that the resulting datasets will help advancing the understanding of honeybee collective intelligence.  


\pagestyle{fancy}

\thispagestyle{empty}

\section{Introduction}\label{introduction}

Honeybees are a popular animal model in biology and have long served as inspiration in computer science. A honeybee colony can itself be seen as a distributed computing system. It manages numerous tasks in parallel with virtuosity. A bee colony adapts to significant environmental variation, it searches for and collects food, feeds its offspring, defends the hive against intruders and regulates temperature and humidity - all without central control. 

In the past, investigating the emergence of the abovementioned collective feats from individual behavior was mostly limited to synthetic approaches \citep{bonabeau1999swarm,beshers2001models,becher2014beehave,dornhaus_benefits_2006}. Analytical approaches through empirical studies have traditionally been limited in various dimensions. Numerous works have focused on the properties and effects of single behaviors, such as the famous waggle dance communication, that serves to direct and optimize the colony's foraging efforts \citep{gruter_steps_2009}. The dance, however, is only one of the many communication channels over which nestmates may receive social information \citep{seeley_wisdom_1995}. Furthermore, due to the time-consuming nature of the biological experiment, virtually all studies were limited in the number of animals under observation, or its duration and sample rate \citep{von_frisch_tanzsprache_1965,scheiner2013standard}.
v
To make observational matters worse, bees are versatile learners. The life of a forager bee typically spans three to four weeks in which she may experience unique environmental features, such as rewards associated with olfactory cues, or locations that may bear perils. Previous work has shown that these memories can modulate a bee's subsequent communication behavior. \citep{gruter_propagation_2006, balbuena_honeybee_2012, gruter_past_2009, gruter_honeybee_2011, goyret_non-random_2005, de_marco_variability_2008, richter_past_1993, nieh_negative_2010}. To fully understand how each of the colony's individual members contribute to the collective, one arguably needs take into account each animals' personal experience during their entire lifetime.

The behaviour of honeybees and other social insects was traditionally studied by manual observation and data collection \citep{von_frisch_tanzsprache_1965, visscher_foraging_1982, seeley_wisdom_1995}. In the last decades, video recording technology was broadly adopted. On video, a larger number of individuals can be studied in detail \citep{beekman_comparing_2004, de_marco_variability_2008,  landgraf_analysis_2011,wario2017automatic}. Effectively studying an entire bee colony over a long period of time, however, requires automation. Software for tracking unmarked animals is increasingly used in the behavioral sciences. Bees often leave the hive and even the young in-hive workers may be lost visually due to frequent occlusions. Because bees can not be identified based on their appearance alone, tracking systems for unmarked animals \citep{khan_rao-blackwellized_2004, landgraf_tracking_2007, veeraraghavan_shape-and-behavior_2008} cannot be used for long-term observations. 

In this paper, we describe a software pipeline for the automatic localization and identification of honeybees using custom binary identification markers. With this pipeline, we processed hundreds of terabytes of raw image data with high accuracy to create a unique dataset containing positions and orientations of all bees in a bee colony spanning several generations of workers. 
\section{State of the Art}\label{previous-work}
Planar markers with binary codes have been shown to be feasible for tracking large groups of insects. A system, previously developed for ants \citep{mersch_tracking_2013} was shown to successfully track 100 bees for two days \citep{blut_automated_2017}. The markers used were originally described as fiducial markers in augmented reality systems \citep{fiala_artag_2005} and rely on spatial derivatives to detect the rectangular outline of a tag. A similar system using flat and rectangular markers for tracking larger insects was also proposed and might be adapted to honeybees \citep{crall_beetag:_2015}. This system binarizes the image globally and searches for rectangular regions representing the corners of the marker. The previous BeesBook vision system \citep{wario_automatic_2015} was tailored to specifically track all animals of small honeybee colonies over their entire lifetime. It uses a round and curved marker and searches for ellipse-shaped edge formations. 

\begin{docfigure}{The BeesBook observation hive. Two high-resolution, low-fps cameras per side are used to locate and identify honeybees. Some behaviours such as the high-frequency waggle dance can not be detected using these cameras. Therefore, an additional camera per side records the hive with a low spatial but high temporal resolution. Figure adapted from \cite{wario_automatic_2015}.}[recsetup-3] 
\docgraph{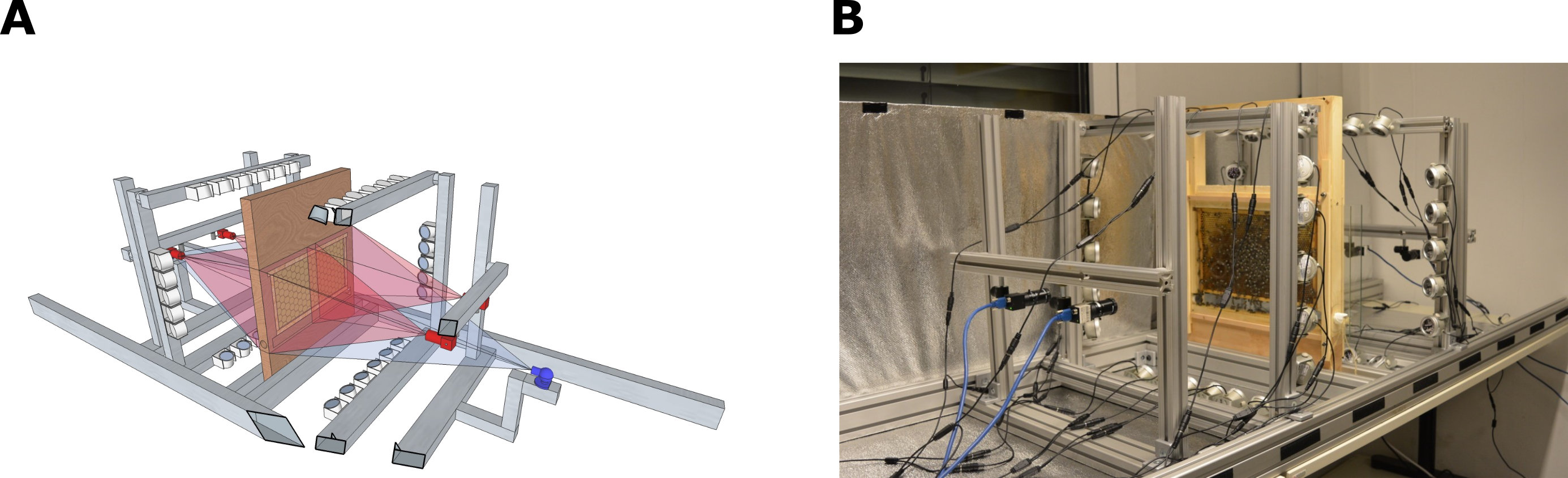}{1}
\end{docfigure}
 
In Wario et al. 2015 we described a pipeline of conventional computer vision steps for detecting and decoding the markers. Although functional, this first prototype was computationally expensive and relied on a powerful supercomputer to process the large amounts of image data we recorded over three summer seasons. Furthermore, the decoding accuracy was drastically dependent on the image quality and respective parameters that had to be tuned for each of the four cameras used in the system.

\begin{docfigure}{Tag design in the \textit{BeesBook} project. The 12 ring segments encode the IDs
        of the animals. The orientation can be inferred from the inner semicircles. Figure adapted from \cite{wario_automatic_2015}.}[tag-design]
    \docgraph{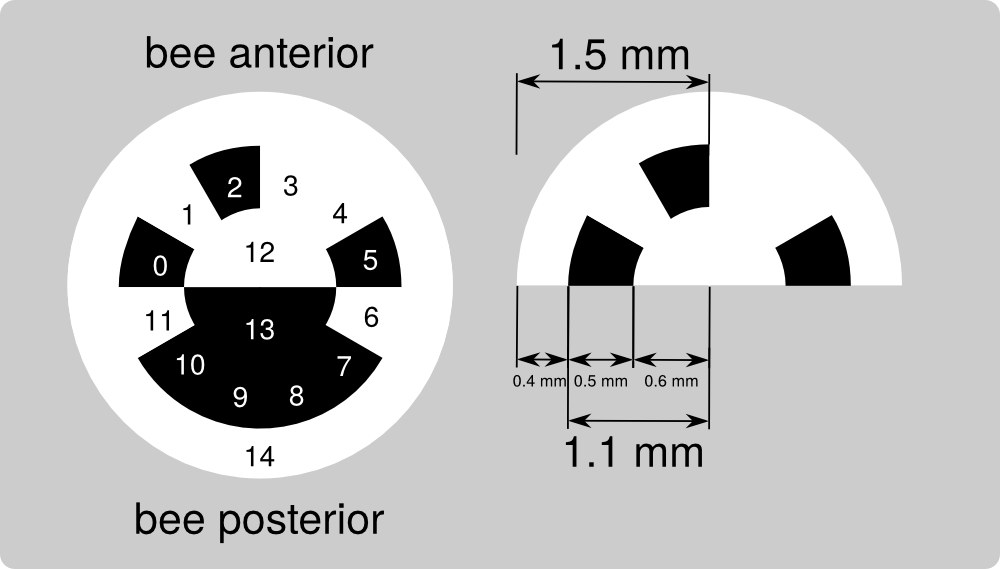}{1} \end{docfigure}

In recent years, deep convolutional neural networks (DCNNs) advanced to the state of the art in many computer vision tasks \citep{krizhevsky_imagenet_2012, he_deep_2015, razavian_cnn_2014}. Modern DCNN architectures such as VGG from the Visual Geometry Group \citep{simonyan_very_2014} and GoogLeNet \citep{szegedy_inception-v4_2016} significantly outperform traditional image recognition techniques in essentially all common image classification benchmarks. Furthermore, deep convolutional neural networks are remarkably successful in object recognition and image segmentation tasks \citep{shelhamer_fully_2016, girshick_rich_2013}. Neural networks have been utilized to detect QR Codes \citep{chou_qr_2015} and to design application specific visual markers \citep{grinchuk_learnable_2016}.

A large amount of available training data is usually necessary to train a modern deep convolutional neural network \citep{sun_revisiting_2017}. 
In our scenario, the target labels are three angles for the marker's spatial rotation and 12 bits that encode the animal's ID. However, extracting these target labels from real image recordings is prohibitively time-consuming. To make things worse, due to their large amount of parameters, DCNNs tend to overfit and fail to generalize when employed in a small-data regime.

Unsupervised and semi-supervised methods can be used in settings where no or only limited amounts of labels, but abundant amounts of unlabeled data are available \citep{bengio_representation_2012}.  Recently, several generative models such as variational autoencoders \citep{kingma_auto-encoding_2013} and generative adversarial networks (GAN) \citep{goodfellow_generative_2014} have been remarkably successful in learning the data distribution of large image datasets. In a recent work, we proposed extending the GAN framework to automatically generate \textit{labeled} data for the honeybee marker decoding scenario \citep{sixt_rendergan:_2016}. Realistic images of bee markers exhibit various kinds of image characteristics, resulting from sensor noise, compression, lighting conditions and motion. 
The RenderGAN learns to reproduce these image characteristics from an unlabeled dataset but ensures that the target labels are preserved. With RenderGAN we generated millions of image-labels-pairs clearing the way for using deep convolutional networks for decoding bee markers (see Figure \ref{fig:random-z-interpolation}). 

\begin{docfigure}{A large dataset of tag images with corresponding labels was automatically generated. Each row shows images of the same tag under different simulated recording conditions. Figure adapted from \cite{sixt_rendergan:_2016}}[random-z-interpolation] 
\docgraph{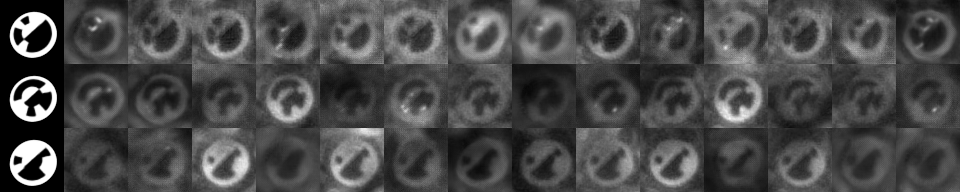}{1}
\end{docfigure}

\section{Methods}
\label{sec:methology}
We identified the following requirements for the proposed machine vision solution. Since in the BeesBook project observations span several weeks, large volumes of image data have to be processed. Hence, the vision system needs to be fast and efficient. Honeybees populate the comb surface densely. The localization component therefore needs high spatial accuracy. Naturally, the detection and decoding accuracies should surpass the baseline described in \cite{wario_automatic_2015}.  

\begin{docfigure}{A visualization of the different processing steps of the pipeline. Beginning from the top left: 1. Original image, 2. Preprocessed image, 3. Localizer, 4. Decoder. The localizer returns the position of the tags on the comb. The decoder evaluates the probability of single bits in the tags to be set. Using the location of the white half in the inner circle, it can also calculate the bees' orientations.}[pipeline_overview] \docgraph{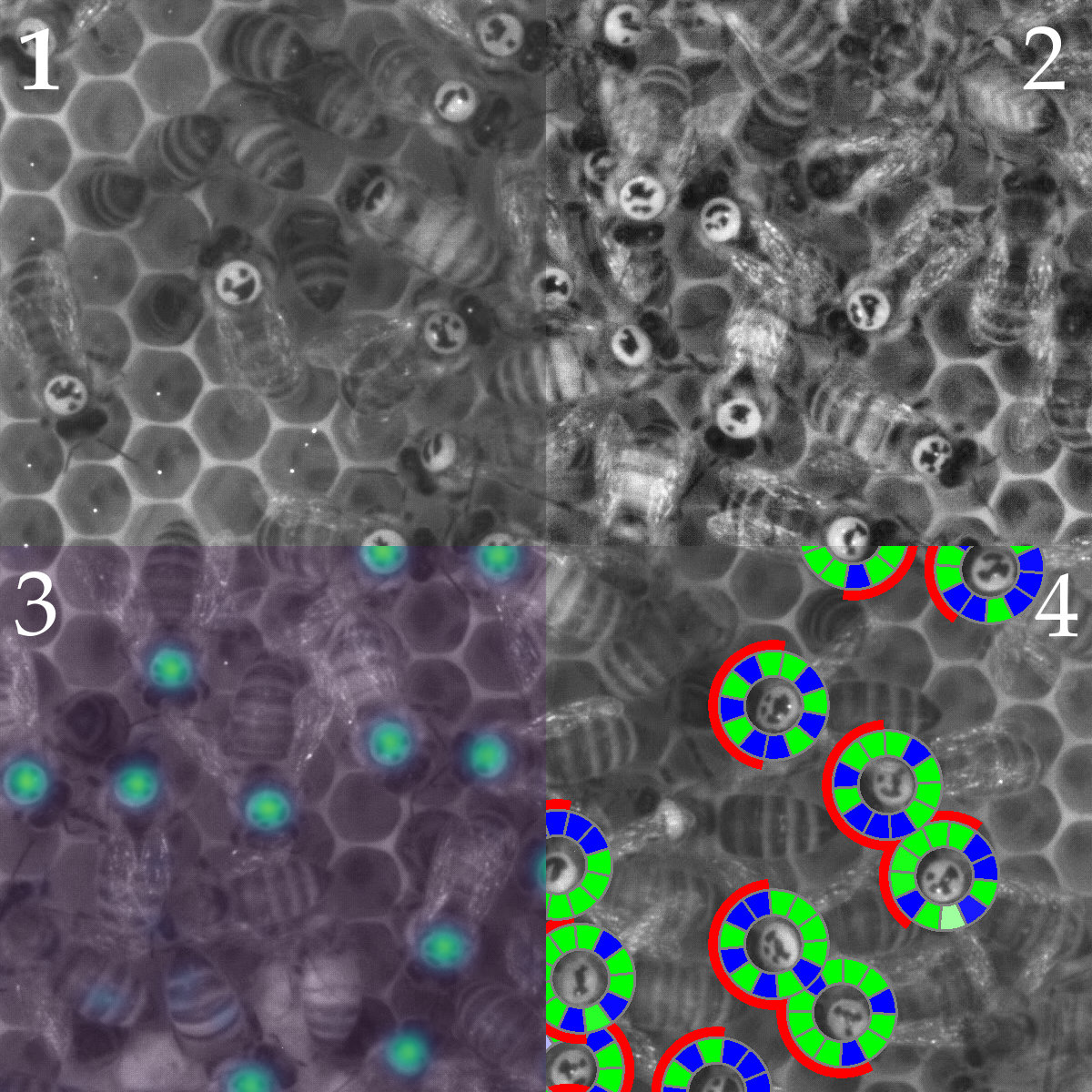}{1} \end{docfigure}

Execution time of convolutional neural networks is proportional to their size. Therefore, we propose using two separate models for marker localization and decoding, respectively. Tag localization does not require a high image resolution and may be a much easier task than decoding the marker. Therefore, a small fully convolutional net is used to localize the markers in a downsampled image. Only image regions containing bee markers are then processed in full resolution by a separate decoder network. This helps reducing processing time significantly. 

\paragraph{Localization}

Positions of honeybee tags were manually annotated in a small subset of the raw BeesBook image data. To this end, the exact position of the center point of each tag was manually labeled in 179 images using a custom GUI. Because only the positions of the tags were required at this stage, this task took us only a few days.

Due to human error, we assume a normal distributed deviation of the center positions and propose using smoothly decaying labels rather than abstract coordinates as follows: Small image patches (128 px width) were randomly sampled from the annotated ground truth dataset as inputs. We approximate the probability that a binary marker is exactly in the center of this image region with the density function of a bivariate normal distribution centered at the nearest true marker position with a fixed variance. The value of the density function at the center position of the image patch is used at the target variable in a regression setting (see Figure \ref{fig:deep-loc-data} for an example).

\begin{docfigure}{A region of interest and the corresponding labels for the localization task. A bright color represents a high probability of the pixel to be in the center of a tag. During training, the probability at the center pixel (inside the red square) is used as the target variable for the image patch.}[deep-loc-data] \docgraph{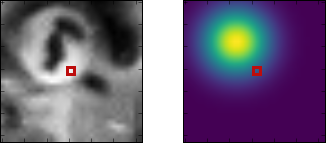}{1} \end{docfigure}

Using this method, a dataset of approximately $300.000$ image patches (based on roughly $10.000$ unique images of tag markers) was generated for the training of a marker localization model. 

Images were preprocessed using the CLAHE \citep{pizer_adaptive_1987} algorithm to reduce the variance in brightness and contrast between different regions of the images, different cameras, and recording seasons. The image regions were then downsampled to a size of 32 px width using bilinear interpolation. Heavy data augmentation (such as random rotations and translations, added gaussian image noise, random elastic deformations, and random brightness and contrast perturbations) were used to prevent overfitting. 

The localization model is a small fully convolutional neural network with three convolutional layers with a kernel size of 5 px (2 px stride) followed by a ReLU activation \citep{glorot_deep_2011}. No padding is applied before the convolutional layers so that the model can be easily applied to full images during inference (and not only to the small regions of interest in the training dataset). Dropout \citep{srivastava_dropout:_2014} is applied following each convolutional layer to reduce overfitting. A final convolutional layer with a kernel size of 1 px (1 px stride) followed by a sigmoid activation computes the target probability for each output pixel.

The network is trained using stochastic gradient descent with momentum for a fixed number of 100 epochs.


During inference, we apply the network to compute a saliency map of the whole input image (after preprocessing and downsampling). Morphological operations combined with a maximum filter are then used to extract the local maxima. Maxima with a saliency below a fixed threshold are discarded. Image patches centered at these local maxima are extracted from the preprocessed images in full resolution and passed on to the next processing stages, \eg the tag decoder or visualization.

\begin{docfigure}{The localization network computes a saliency map of potential tag locations. After post-processing the saliency map with morphological operations and thresholding, high probabilty candidates are extracted using a maximum filter.}[deep-loc-saliency] 
\docgraph{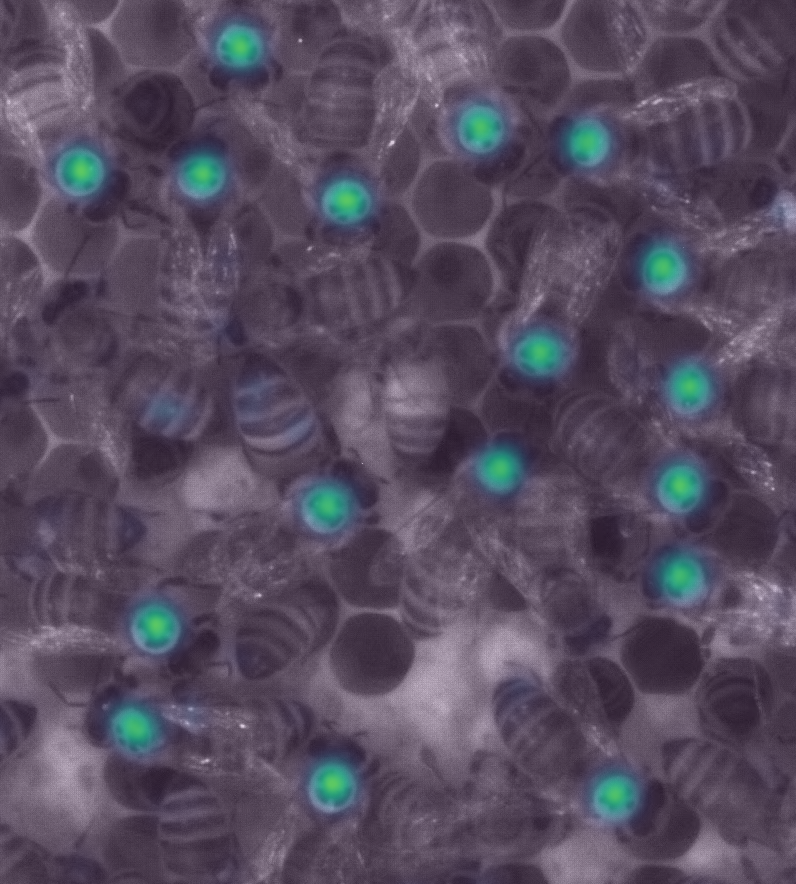}{2}
\docgraph{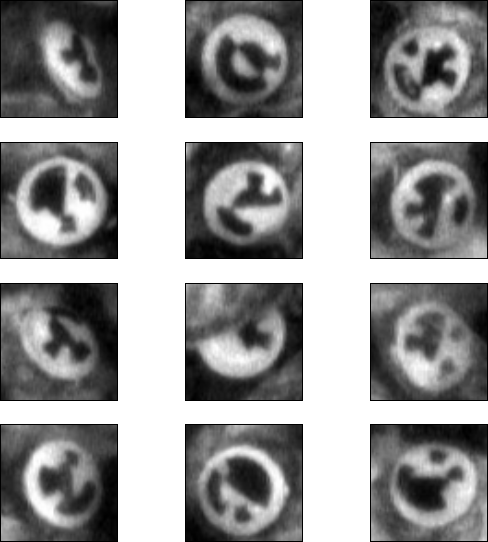}{2} 
\end{docfigure}

\paragraph{Decoding}

While creating a labeled dataset for the localization task was manageable, it was unreasonably time-consuming for the decoding task. All members of our group used a custom GUI to visually match a three-dimensional grid onto all visible markers in a single camera recording. We obtained a labeled dataset of about 2000 marker instances in about one week time. This dataset served as evaluation set for the final system and not to train the convnet.

\begin{docfigure}{Localizer and Decoder architecture. A small network (A) localizes markers on a downsampled input. Image patches centered at the positions of the markers are processed in full resolution by the larger decoder network (B).}[deep-dec-data]
    
    \noindent\resizebox{\textwidth}{!}{
	\begin{tikzpicture}
		\draw[use as bounding box, transparent] (-1.8,-2.4) rectangle (23, 4.2);
		\newcommand{\networkLayer}[6]{
			\def\a{#1}
			\def\b{0.02}
			\def\c{#2}
			\def\t{#3}
			\ifthenelse {\equal{#6} {}} {\def\y{0}} {\def\y{#6}}
			\draw[line width=0.25mm](\c+\t,0,0) -- (\c+\t,\a,0) -- (\t,\a,0);                                                      
			\draw[line width=0.25mm](\t,0,\a) -- (\c+\t,0,\a) node[midway,below] {#5} -- (\c+\t,\a,\a) -- (\t,\a,\a) -- (\t,0,\a);
			\draw[line width=0.25mm](\c+\t,0,0) -- (\c+\t,0,\a);
			\draw[line width=0.25mm](\c+\t,\a,0) -- (\c+\t,\a,\a);
			\draw[line width=0.25mm](\t,\a,0) -- (\t,\a,\a);
			\filldraw[#4] (\t+\b,\b,\a) -- (\c+\t-\b,\b,\a) -- (\c+\t-\b,\a-\b,\a) -- (\t+\b,\a-\b,\a) -- (\t+\b,\b,\a);
			\filldraw[#4] (\t+\b,\a,\a-\b) -- (\c+\t-\b,\a,\a-\b) -- (\c+\t-\b,\a,\b) -- (\t+\b,\a,\b);
			\ifthenelse {\equal{#4} {}}
			{}
			{\filldraw[#4] (\c+\t,\b,\a-\b) -- (\c+\t,\b,\b) -- (\c+\t,\a-\b,\b) -- (\c+\t,\a-\b,\a-\b);}
		}
        
        \node[] at (0,-2) {\Large \textbf{(A) Localizer}};
        
        \networkLayer{1.5}{0.0}{-0.5}{color=gray!80}{}{}

		\networkLayer{1.5}{0.2}{0}{color=white}{5x5x[32,64,128]}{}
		\networkLayer{0.8}{0.4}{0.5}{color=white}{}{}
		\networkLayer{0.4}{0.6}{1.0}{color=white}{}{}
        \networkLayer{0.2}{0.0}{1.75}{color=gray!80}{}{}

		\networkLayer{4.0}{0.0}{4.0}{color=gray!80}{}{}

		\networkLayer{4.0}{0.1}{4.5}{color=white}{}{}
		\networkLayer{4.0}{0.1}{4.7}{color=white}{}{}
		\networkLayer{4.0}{0.1}{5.0}{color=white}{}{}
		\networkLayer{4.0}{0.1}{5.2}{color=white}{3x3x16}{}
		\networkLayer{4.0}{0.1}{5.5}{color=white}{}{}
		\networkLayer{4.0}{0.1}{5.7}{color=white}{}{}

		\networkLayer{2.0}{0.2}{6.0}{color=white}{}{}
		\networkLayer{2.0}{0.2}{6.3}{color=white}{}{}
		\networkLayer{2.0}{0.2}{6.7}{color=white}{}{}
		\networkLayer{2.0}{0.2}{7.0}{color=white}{}{}
		\networkLayer{2.0}{0.2}{7.4}{color=white}{3x3x32}{}
		\networkLayer{2.0}{0.2}{7.7}{color=white}{}{}
		\networkLayer{2.0}{0.2}{8.1}{color=white}{}{}
		\networkLayer{2.0}{0.2}{8.4}{color=white}{}{}

		\networkLayer{1.0}{0.4}{8.8}{color=white}{}{}
		\networkLayer{1.0}{0.4}{9.3}{color=white}{}{}
		\networkLayer{1.0}{0.4}{9.9}{color=white}{}{}
		\networkLayer{1.0}{0.4}{10.4}{color=white}{}{}
		\networkLayer{1.0}{0.4}{11.0}{color=white}{}{}
		\networkLayer{1.0}{0.4}{11.5}{color=white}{}{}
		\networkLayer{1.0}{0.4}{12.1}{color=white}{3x3x64}{}
		\networkLayer{1.0}{0.4}{12.6}{color=white}{}{}
		\networkLayer{1.0}{0.4}{13.2}{color=white}{}{}
		\networkLayer{1.0}{0.4}{13.7}{color=white}{}{}
		\networkLayer{1.0}{0.4}{14.3}{color=white}{}{}
		\networkLayer{1.0}{0.4}{14.8}{color=white}{}{}
        
        \node[] at (12,-2) {\Large \textbf{(B) Decoder}};

		\networkLayer{0.5}{0.8}{15.4}{color=white}{}{}
		\networkLayer{0.5}{0.8}{16.3}{color=white}{}{}
		\networkLayer{0.5}{0.8}{17.3}{color=white}{}{}
		\networkLayer{0.5}{0.8}{18.2}{color=white}{3x3x128}{}
		\networkLayer{0.5}{0.8}{19.2}{color=white}{}{}
		\networkLayer{0.5}{0.8}{20.1}{color=white}{}{}
        
        \networkLayer{0.2}{1.6}{21.1}{color=gray!80}{4x4x256}{}

	\end{tikzpicture}
	}
\end{docfigure}

Due to the complexity of the labels and the time-consuming nature of manual labeling, we developed a method to generate realistic images of markers that correctly display a given label. This work combines a typical GAN architecture \citep{goodfellow_generative_2014} with a 3D model of the bee markers. The GAN's generator consists of a number of image augmentation functions that sequentially add image characteristics such as background, blur, noise and lighting to synthetic images originating from the 3D model. The parameters of these augmentations are learned through adversarial training \citep{goodfellow_generative_2014} leveraging the previously mentioned database of marker images. The approach, called RenderGAN, allowed generating marker images with arbitrary combinations of spatial rotations and bit configurations. It was used to create a large dataset of 5 millions of labeled marker images for the training of a decoder model. See~\cite{sixt_rendergan:_2016} for more details.

A large convolutional neural network based on the ResNet architecture \citep{he_deep_2015} is used to decode the localized tags. Image patches containing markers with a size of 64 px are fed to an initial convolutional layer with a kernel size of 3 px (1 px stride) followed by a Batch Normalization \citep{ioffe_batch_2015} layer and an ELU \citep{clevert_fast_2015} activation. The data is then processed following the 34-layer architecture described in \cite{he_deep_2015}, but starting with only 16 filters in the first block to reduce the total numbers of parameters. The resulting representations are then individually processed by two fully connected layers with 256 filters each and followed by an ELU activation. The output from the first fully connected layer is finally used to compute the probabilities for each bit to be set using a sigmoid activation. The second fully connected layer is used to compute the spatial rotations of the marker (represented as vectors on the unit circle) and a full resolution offset for the exact center position of the tag. The bit probability outputs are optimized using a cross entropy loss while the mean squared error is used for the other outputs. 

Potential decoding errors can still be corrected in later processing stages, \eg during temporal
tracking of the detections over time. To faciliate this, we store the raw probabilites for each bit
instead of the thresholded predictions for each detection. Furthermore, we define a confidence
measure for each prediction as:

\begin{equation}
c(b)=\frac{\prod_{i=0}^{12}2\cdot|0.5-p(b_{i})|}{12}
\end{equation}

A high confidence measure $c(b)$ signifies that the decoder assigns a high probability to each of
its predictions for the individual bits $p(b_{i})$ which means that the decoding of the ID is
likely to be correct. This measure can also be used to discard low confidence predictions in analyses
where a high precision of the decoded IDs is more important than a high recall of detected bees.

\paragraph{Data storage}

While all data in the BeesBook project is ultimately stored in a database that can be used and extended easily by other team members, an intermediate data format was used to store the outputs of the pipeline due to the following reasons:
Concurrently writing to a central database is difficult to achieve because of the peculiarities of a supercomputer's job queueing system and creates a potential IO bottleneck (\ie the workers can't continue to process data because they have to wait for the database to store their previous results). Furthermore, the pipeline should work without any central coordination, \ie it should be possible for any worker to process its local subset of the data without any communication to another process. Lastly, writing to the distributed filesystem on a supercomputer can be slow and therefore the size of the outputs should be as small as possible.

For these reasons, the results of the pipeline were serialized and saved using the Cap'n Proto library \citep{capnproto}. After processing, the results of the individual workers were collected and sorted by date in a simple file system hierarchy.

\section{Results}\label{results}

We evaluated the performance of the two networks and compared them with our previous computer vision pipeline \citep{wario_automatic_2015} and the bee tracking systems described in \cite{blut_automated_2017} and \cite{gernat_automatic-monitoring-2018}. We also include the results of our approach when combined with temporal tracking as described in \cite{boenisch_feature_2017}.

\paragraph{Localizer}
The localizer convnet was trained to predict the saliency label as described in the previous section. During training, the model learns to minimize the binary cross-entropy between its outputs and the training labels. While this metric has proven to
be very effective for training, it is less meaningful as a performance metric for detecting bees. 
We therefore calculate recall (the ratio of detected markers and the number of existing markers in the same image) and precision (percentage of regions that correctly contained a marker). Both metrics depend on the threshold value used to discard local maxima with a low probability. A threshold of $0.6$ was determined empirically and used in the evaluation and data processing.

The new localizer outperforms the localization stage of the old computer vision pipeline in terms of runtime, recall, and precision without the need for hyperparameter tuning. The model can be applied without performance loss on images from all recording seasons and is able to recognize tags on images that are very different from the BeesBook observation hive recordings, for example tags on a white table photographed with a smartphone camera. Qualitative inspection of error cases suggests that the model sometimes misses tags which are very close to the image border (where the image sharpness is lower) and tags outside of the main surface area of the comb (where tags are usually farther away from the camera).

\begin{docfigure}{Comparison of recall and precision rates for localizing honeybee markers. The system we propose achieves best recall rates, i.e. our localizer misses the least amount of markers. In comparison to our previous system, we improved precision rates by $11 \%$. In \cite{blut_automated_2017} and \cite{gernat_automatic-monitoring-2018},  precision rates for localizing markers were not reported. }[stats-localizer]
\begin{tabular}{|c|c|c|}
\hline
 & Recall & Precision\tabularnewline
\hline
\cite{wario_automatic_2015} & $94\pm4\%$ & $88\pm4\%$\tabularnewline
\hline
\cite{blut_automated_2017} & $90.8\%$ $-$ $98.2\%$ * & $-$\tabularnewline 
\hline
\cite{gernat_automatic-monitoring-2018} & $87\pm2\%$ & $-$\tabularnewline
\hline
\textbf{This work} & $98.3\%$ & $99.4\%$\tabularnewline
\hline
\end{tabular}

{* \footnotesize Moving bees vs.\ resting bees}
\end{docfigure}

\paragraph{Decoder}
The decoder model not only predicts the values of the individual bits on the tag, but also all three spatial rotations. During training, the decoder learns to minimize the binary cross-entropy between the true values of the individual bits and its predictions and the mean squared error between the true orientations and its predictions.

We evaluate the mean hamming distance (the average number of bits decoded incorrectly) and the decoding accuracy (the number of tags that were decoded without any error).

Even though we do not use ECC, the decoder is able to decode $87.8\%$ of the tags without errors. If we combine predictions from this DCNN with temporal tracking, the accuracy of the assigned IDs improves to $98.1\%$ compared to $66\%$ when using our computer vision pipeline \citep{boenisch_feature_2017}.

\begin{docfigure}{Comparison of marker decoding performance. Our systems do not use error correction schemes and may therefor exhibit flipped bits. The mean hamming distance (the expected number of flipped bits) was improved significantly. The proportion of correctly decoded markers (\ie zero flipped bits) was improved by $21 \%$. The system proposed in \cite{gernat_automatic-monitoring-2018} surpasses our raw result by $11 \%$. Applying a post-processing step to link corresponding detections to motion paths \citep{boenisch_feature_2017} allows us to match this decoding accuracy. Because of the improved performance, we can process our data in realtime on consumer hardware.}[stats-decoder]
\begin{tabular}{|c|c|c|c|}
\hline
 & MHD & Accuracy & Time per tag (ms) \tabularnewline
\hline
\cite{wario_automatic_2015} & $1.08$ & $66\%$ & $177.54$ \tabularnewline
\hline
\cite{gernat_automatic-monitoring-2018} & $-$ & $98.58\%$ & $-$ \tabularnewline
\hline
\textbf{This work (w / wo tracking)} & $0.42 / 0.08$ & $87.8\% / 98.1\%$ & $1.43 / 2.01$ \tabularnewline
\hline
\end{tabular}
\end{docfigure}

\begin{docfigure}{Confidence measure of decoder vs accuracy. Depending on the analysis, either a high recall or a high precision may be preferable. At maximum recall, the model can correctly decode approximately $88\%$ of all detections. A very high decoding accuracy of $99.14\%$ can be achieved without any temporal tracking by discarding the $40\%$ of the detections with the lowest confidence score.}[deep-conf-acc] \docgraph{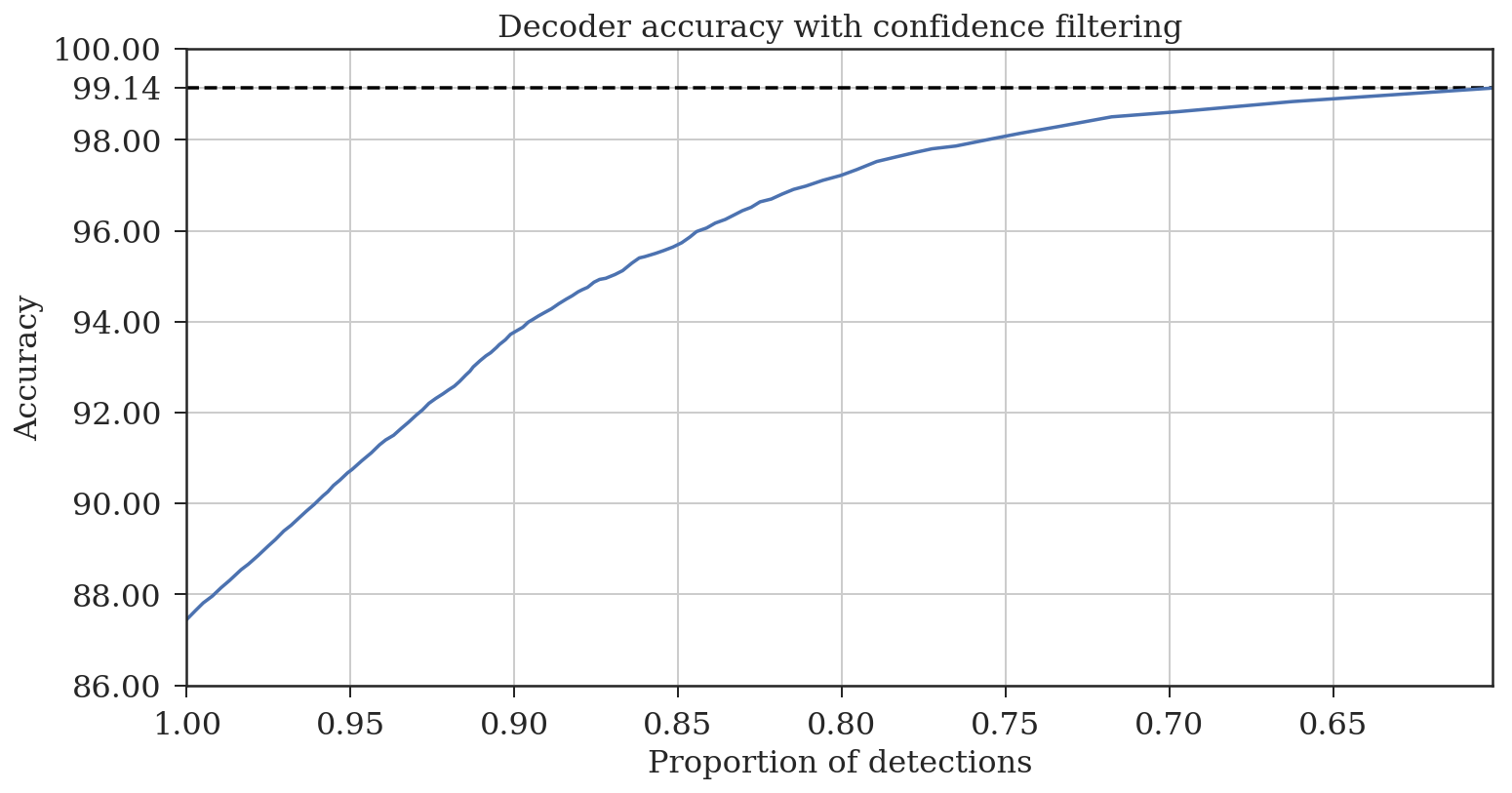}{1}
\end{docfigure}

\begin{docfigure}{Visualization of the pipeline results on an entire image. Almost all tags are correctly localized and decoded with a high confidence (a high saturation of the colors in the overlay represents a high confidence of the decoder).}[deep-vis]
    \docgraph{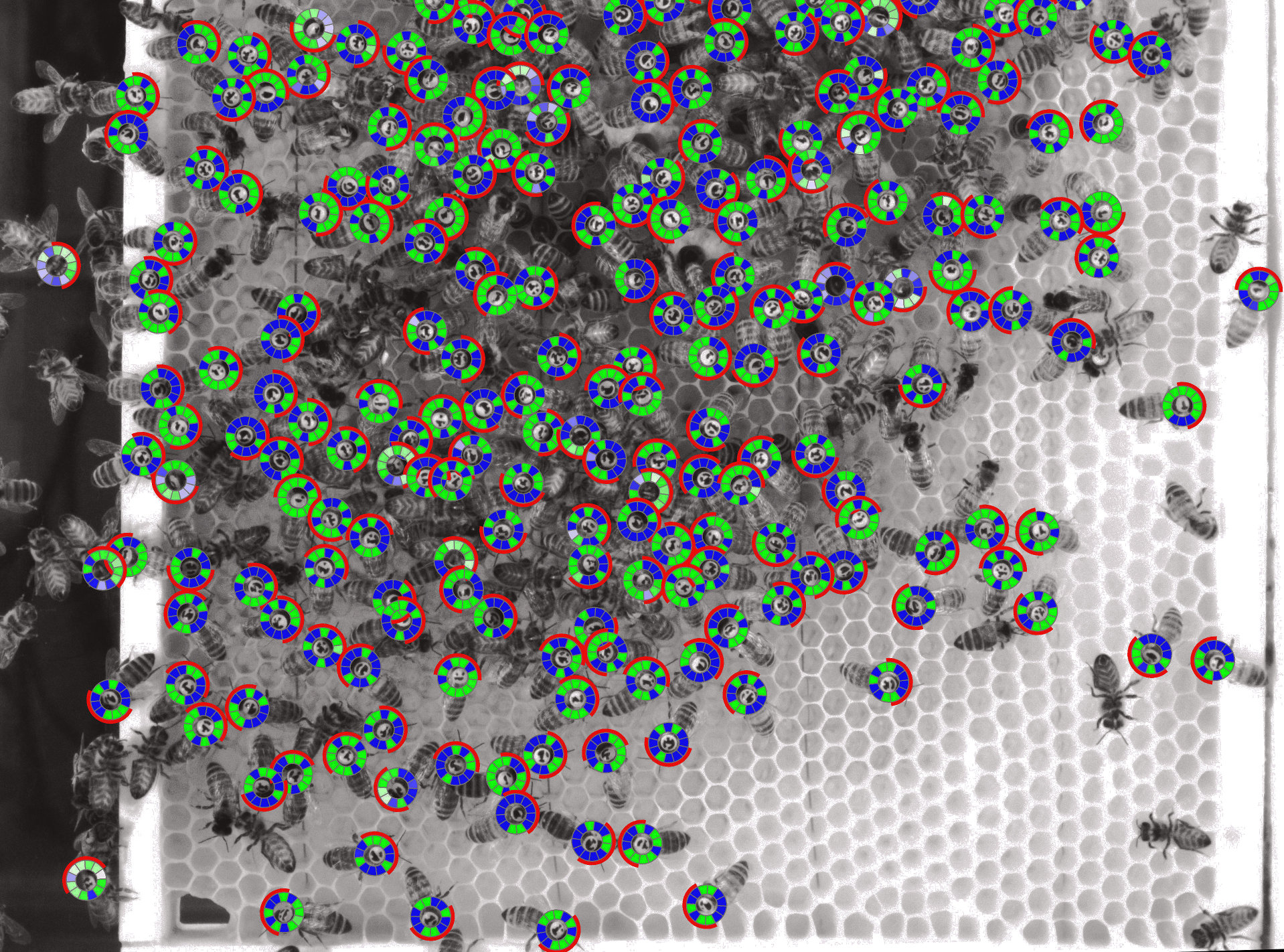}{1}
\end{docfigure}

\paragraph{Resulting datasets}

All data of the recording seasons 2015 and 2016 was processed using the new deep learning pipeline on a Cray X30 supercomputer. The datasets are now available for further studies of the honeybee's social behavior. Temporal tracking of the detections over time further improves the quality of the datasets \citep{boenisch_feature_2017}.

In total, $3.614.742.669$ honeybees were detected on $67.972.617$ images for the 2015 recording season ($\approx 53$ detections per image). In 2016, there were $6.331.078.577$ detections in $59.680.181$ images ($\approx 106$ detections per image). The serialized results are 274 GB in size for 2015 and 476 GB for 2016.

Jobs on the HLRN Cray supercomputer are billed in NPL (north-german parallel computer work units). We used Intel Xeon Haswell compute nodes for which $0.1$ NPL are billed per core hour.  All in all, processing both datasets was billed at $39.896$ NPL which corresponds to roughly $398.954$ core hours in total. Under optimal circumstances with no IO contention, the data could therefore be processed in roughly one week using 1200 CPUs (100 mpp2 compute nodes on the Cray X30 supercomputer). Please note that the data can also be processed in realtime using a consumer GPU (Geforce GTX 1080 Ti) instead of CPUs. In contrast, processing the data of a single recording season using computer vision pipeline described in \cite{wario_automatic_2015} was billed with more than $200.000$ NPL.
\section{Discussion}\label{discussion}

We have presented a new method for the automatic localization and decoding of honeybee markers using deep convolutional neural networks. We are able to identify virtually all individuals in a honeybee colony over many weeks. Tracking individual honeybees with barcode-like markers has only recently been proposed by several groups \citep{wario_automatic_2015,crall_beetag:_2015,blut_automated_2017,gernat_automatic-monitoring-2018} and will likely allow unprecedented perspectives on the complex interplay between all colony members. While all other systems rely on planar rectangular markers, we decided for an unconventional curved and round marker design. This decision was motivated by our interest in long-term observations and preliminary experiments showing that planar markers endure less mechanical stress and might fall off after only a few days. 

While the image processing steps to detect and decode rectangular markers are well researched, we found that conventional, i.e. non-neural computer vision algorithms were less suitable for our custom markers, especially regarding the runtime. Using deep convolutional neural networks as described in this paper, we reached similar, or better detection and decoding accuracy and significantly improved runtime performance compared to comparable approaches using a more traditional marker design \citep{blut_automated_2017, gernat_automatic-monitoring-2018}.

The performance comparison in section \ref{results} might be arguable due to differences in methodology and data used. While our system does not use any error correction, we discriminate between detection and decoding. The system used in \citep{blut_automated_2017} might return a detection only if it could be decoded correctly. In this case, the detection accuracy may actually indicate the decoding accuracy, i.e. the proportion of correctly decoded markers. The authors report variable detection performance (between $90.8 \%$ and $98.2 \%$) and attribute lower performance to situations in which many bees are in motion. The general level of activity in a bee colony depends on many factors. Under natural conditions there may be a fair amount of motion throughout the day and we therefore expect the decoding accuracy to vary closer to the lower bound. Observations over night may result in decoding accuracies near the upper bound. 

In \citep{gernat_automatic-monitoring-2018} the authors report $87 \%$ mean detection accuracy which is lowest among the systems. It remains unclear how many false positives these detections contain. The decoding accuracy for this system ($98.58 \%$) is highest among the systems. It remains unknown how many of the potential false detections can be identified in the decoding step. In the best case, this system, hence, has an overall accuracy of $85.7 \%$. The BeesBook system described here yields a similar combined detection and decoding accuracy of $98.3\% \cdot 87.8\% = 86.3\%$. The system described in \citep{crall_beetag:_2015} was validated without manually generated ground truth data and therefore was excluded from the comparison. 

In \cite{boenisch_feature_2017} we propose an additional postprocessing step to significantly increase the decoding accuracy. Linking corresponding detections through time (tracking) and averaging bit probabilities of detections within a path increased the combined detection and decoding accuracy to $96\%$. 

Depending on the research question, different properties may be relevant for the end-user of such tracking systems. The BeesBook system offers a very low rate of false detections, a high decoding accuracy and longevity of the markers with relatively inexpensive hardware components. We think this system is suited to tackle a broad range of research questions and we invite researchers to test it. 

Our method relies on two machine learning models, the localization and decoding networks. Because of this design decision, we can adapt our software easily to changes in the recording setup. Significant modifications, for a example a new tag design, requires retraining both models, which in most cases may be much easier than redesigning a pipeline of conventional computer vision steps. In our system, the following steps have to be taken: A new training dataset for the localization task has to be created manually. A new localization model can then be trained. The new model can then be used to generate training data for the RenderGAN \citep{sixt_rendergan:_2016}. Finally, data generated by the RenderGAN can then be used to train a new decoder model. Thus, the only manual labor is clicking sample markers images for the localization training dataset. All remaining steps are automated. In many cases, retraining our models may not even be necessary. Exploratory experiments with cameras used in field experiments suggest that the convolutional neuronal networks are surprisingly invariant to changes in illumination, lens distortion and image quality.

A significant result is the 100-fold speedup we achieved in comparison to our previous prototype. Using a consumer-grade graphics card (Geforce GTX 1080 Ti) we obtained realtime performance with a 3 Hz recording framerate and approximately 800 individuals in the colony. In the past, we have recorded full image datasets spanning up to nine weeks of continuous experiment. In total, we currently store three datasets (63 days and ~45 TB each) on tape drives granted by our project parter, the North-German Supercomputing Alliance (HLRN). Sample image data is available upon request. A trajectory dataset of all animals over three continuous days is available online \citep{boenisch_beesbook_dataset_2018}.

\bibliography{main}{}
\bibliographystyle{agsm}

\end{document}